\documentclass[letterpaper, 10 pt, conference]{ieeeconf}  

\IEEEoverridecommandlockouts                              

\usepackage{graphics} 
\usepackage{graphicx}
\usepackage{epsfig} 
\usepackage{amsmath} 
\usepackage{amssymb}  
 \usepackage{algorithm}
\usepackage{algpseudocode}
\algnewcommand\AAND{\textbf{ and }}
\algnewcommand\Or{\textbf{ or }}
\usepackage{color}
\usepackage{citesort}
\usepackage{flushend}

\usepackage{textcomp}

\DeclareMathAlphabet{\pazocal}{OMS}{zplm}{m}{n}

\newcommand{\Ws}{\pazocal{W}}

\newcommand{\Bs}{\pazocal{B}}

\newcommand{\Vs}{\pazocal{V}}

\newcommand{\Ps}{\pazocal{P}}

\newcommand{\Ms}{\pazocal{M}}

\newcommand*{\vertbar}{\rule[-1ex]{0.5pt}{3.25ex}}
\newcommand*{\horzbar}{\rule[.15ex]{8.1cm}{0.25pt}}

\usepackage{array}
\newcolumntype{C}[1]{>{\centering\arraybackslash}p{#1}}
\newcolumntype{M}[1]{>{\raggedright\arraybackslash}p{#1}}

\usepackage{array} 
\newcolumntype{L}[1]{>{\raggedright\let\newline\\\arraybackslash\hspace{0pt}}m{#1}}	
\newcolumntype{S}[1]{>{\centering\let\newline\\\arraybackslash\hspace{0pt}}m{#1}}
\newcolumntype{R}[1]{>{\raggedleft\let\newline\\\arraybackslash\hspace{0pt}}m{#1}}

\title{\LARGE \bf
Visual--Inertial Odometry--enhanced Geometrically Stable ICP for Mapping Applications using Aerial Robots
}

\author{Tung Dang, Shehryar Khattak, Christos Papachristos, Kostas Alexis
\thanks{This material is based upon work supported by the Department of Energy under Award Number [DE-EM0004478].}
\thanks{The authors are with the Autonomous Robots Lab, University of Nevada, Reno, 1664 N. Virginia, 89557, Reno, NV, USA
        {\tt\small tung.dang@nevada.unr.edu}}%
}

\begin{document}

\maketitle
\thispagestyle{empty}
\pagestyle{empty}

\begin{abstract}

This paper presents a visual--inertial odometry--enhanced geometrically stable Iterative Closest Point (ICP) algorithm for accurate mapping using aerial robots. The proposed method employs a visual--inertial odometry framework in order to provide robust priors to the ICP step and calculate the overlap among point clouds derived from an onboard time--of--flight depth sensor. Within the overlapping parts of the point clouds, the method samples points such that the distribution of normals among them is as large as possible. As different geometries and sensor trajectories will influence the performance of the alignment process, evaluation of the expected geometric stability of the ICP step is conducted. It is only when this test is successful that the matching, outlier rejection, and minimization of the error metric ICP steps are conducted and the new relative translation and rotational components are estimated, otherwise the system relies on the visual--inertial odometry transformation estimates. The proposed strategy was evaluated within handheld, automated and fully autonomous exploration and mapping missions using a small aerial robot and was shown to provide robust results of superior quality at an affordable increase of the computational load. 

\end{abstract}

\section{INTRODUCTION}\label{sec:intro}
Aerial robots experience an unprecedented process of wide integration in many critical domains. Within those, the applications related with mapping are among those with the most major impact. Relevant examples include infrastructure inspection~\cite{RHEM_ICRA_2017,bircher2016receding,BABOOMS_ICRA_15,NBVP_ICRA_16,yoder2016autonomous,karrer2016real,SIP_AURO_2015,APST_MSC_2015,DABS_ICRA_14,balta2017integrated,papachristos2016augmented}, precision agriculture~\cite{Zermas2015}, and surveying~\cite{nex2014uav}. In these domains, the ability of robots to perform high quality mapping and conduct the mission autonomously is the key. Furthermore, it is not rare that such tasks have to be conducted in GPS--denied environments (e.g. indoors or very close to structure), while a prior map is not available.

A large body of work exists aiming to address the problem of GPS--denied localization and mapping for aerial robots. Examples include vision-- and visual--inertial~\cite{bloesch2015robust,leutenegger2015keyframe}, LiDAR--based~\cite{zhang2014loam} and other depth range sensors--based~\cite{whelan2012kintinuous} methods. Within those, each sensing modality presents specific advantages and limitations when it comes to the robustness of the pose estimation process, sensor data ill--conditioning for different environments and viewpoints, the map quality, geometric density and more. 

%
\begin{figure}[h!]
\centering
  \includegraphics[width=0.99\columnwidth]{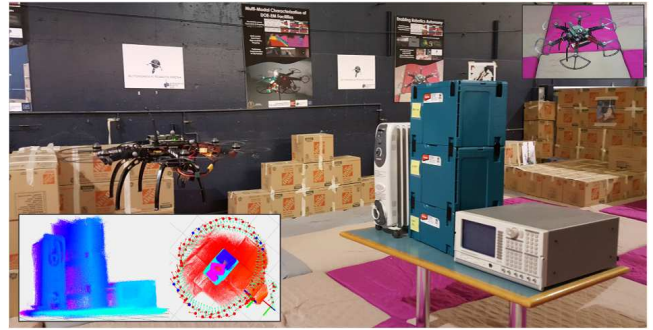}
\caption{An aerial robotic mapping experiment employing the proposed visual--inertial odometry--enhanced geometrically stable ICP algorithm. }
\label{fig:intro}
\end{figure}
%

The approach described in this paper contributes further into the direction of multi--modal mapping for increased $3\textrm{D}$ reconstruction quality. Towards that goal it proposes a Visual--Inertial (VI) odometry--enhanced strategy for a geometrically stable Iterative Closest Point (ICP) algorithm using Time--of--Flight (ToF) short--range depth sensors. In particular, pose estimates provided through a VI odometry framework~\cite{bloesch2015robust} are used a) to calculate the overlap of two consecutive point clouds, and b) provide a transformation prior to the ICP step. When the overlap is sufficient, a geometric stability check will be conducted to assess the expectation of deriving an accurate estimate of the transformation between the two consecutive point clouds using ICP and avoid optimization based on ill--conditioned data that cannot constrain the solution. As long as this test is successful, the framework will proceed to the ICP optimization step, otherwise the method relies on the VI odometry--provided transformations. Within the ICP step, the method of point sampling such that the distribution of normals among them is as large as possible is employed. Then a point--to--plane error metric is defined and minimized~\cite{rusinkiewicz2001efficient,rusinkiewicz2001efficient,rusu20113d,Besl92,turk1994zippered,masuda1996registration,pomerleau2013comparing,mitra2004registration}.

The proposed strategy was thoroughly evaluated in both handheld tests and experiments involving an aerial robot performing automated and autonomous inspection flights. Figure~\ref{fig:intro} presents an instance of these experiments. For both the handheld and aerial robotic experiments, the perception unit consists of a software--synchronized stereo visual--inertial module alongside a pendrive--sized $3\textrm{D}$ ToF depth sensor. Handheld tests are conducted to preliminary evaluate the methodology. The aerial robotic tests involve the use of a platform capable of running all perception, control and planning algorithms onboard and in real-time. Autonomy--wise, the crown of this work is the integration of the proposed algorithm with an uncertainty--aware receding horizon exploration and mapping planner~\cite{RHEM_ICRA_2017} that ensures full autonomy in environments for which no prior knowledge exists. Prior to that an inspection along predefined waypoints in order to map a known structure is conducted. The relevant datasets are also released online.

The remainder of this paper is structured as follows. Section~\ref{sec:approach} details the proposed algorithm and methodological steps. Section~\ref{sec:planner} summarizes the employed autonomous exploration planner, followed by the experimental studies in Section~\ref{sec:experiments}. Finally, conclusions are drawn in Section~\ref{sec:concl}.

\section{PROPOSED APPROACH}\label{sec:approach}
\begin{figure}[h!]
\centering
  \includegraphics[width=0.99\columnwidth]{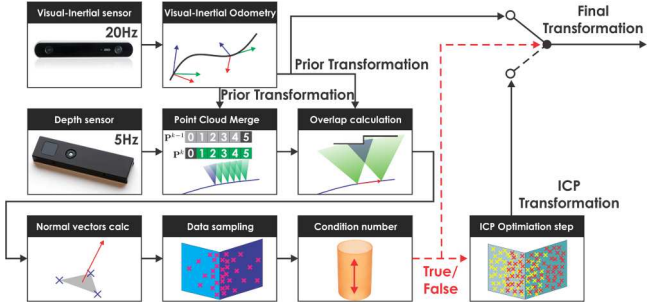}
\caption{The key steps involved in the process of the proposed visual--inertial odometry--enhanced geometrically stable ICP. }
\label{fig:viicip_architecture}
\end{figure}

The proposed approach combines the robust, yet not always high--accuracy, odometry derived from a visual--inertial sensor with the improved mapping properties of ICP algorithms efficiently designed and operating based on the data of a ToF depth sensor such as the PMD Picoflexx. At first, the data from the VI--sensing modalities are utilized within the robust visual--inertial odometry framework described in~\cite{bloesch2015robust} and through that robot localization information becomes available. Given that the depth sensor is calibrated against the VI--sensor, the aforementioned localization information can be used as a prior for the ICP algorithm. Furthermore, it can be exploited to identify the overlapping subset of two consecutive point clouds of the depth sensor and perform ICP only within this selective subset of the data only. When sufficient overlap is calculated, the proposed approach will further evaluate the expected geometric stability of ICP. Only if this is positively assessed, the method will proceed with the ICP step, otherwise the system will purely rely on VI odometry. In addition, since priors for the ICP optimization are available, we afford to run the ICP step even at low frame rates and allow it for more computational time in order to arrive to a further optimized result. The overall architecture is visualized in Figure~\ref{fig:viicip_architecture}. 

%
%

The employed VI odometry pipeline is summarized in Section~\ref{subs:vislam}, followed by the description of the method to use this odometry as a prior for a geometrically stable ICP algorithm running onboard small robotic systems.

\subsection{Visual--Inertial Localization}\label{subs:vislam}
For the purposes of robot navigation as well as reliable prior and last--resort for mapping, a visual--inertial odometry framework is employed due to the enhanced robustness such methods provide. In particular, the open--source Robust Visual Inertial Odometry (ROVIO) is utilized~\cite{bloesch2015robust}. Within this paper, a necessarily brief summary will be provided as the proposed method makes extensive use of it. ROVIO closely couples the tracking of multilevel image patches with the Extended Kalman Filter (EKF) through the direct use of image intensity errors to derive the filter innovation term and uses a QR--decomposition to reduce the dimensionality of the error terms, therefore keeping the Kalman filter update step computationally tractable. Its formulation is robocentric, therefore the landmarks are estimated with respect to the camera pose. The estimated landmarks are decomposed into a bearing vector, as well as a depth parametrization. The Inertial Measurement Unit (IMU) fixed coordinate frame ($\Bs$), the camera fixed frame ($\Vs$) and the inertial frame $\Ws$ are considered and the employed state vector with dimension $l$ and associated covariance matrix $\mathbf{\Sigma}_l$ is:

%
\begin{figure}[h!]
\centering
  \includegraphics[width=0.85\columnwidth]{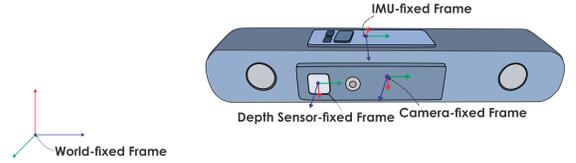}
\caption{Employed coordinate frames.  }
\label{fig:coordframes}
\end{figure}
%

\vspace{-1.5ex}
\begin{eqnarray}\label{eq:roviostate}
 \mathbf{x} = [ \underbrace{\overbrace{\mathbf{r}~\mathbf{q}}^\text{pose, $l_p$}\boldsymbol{\upsilon}~\mathbf{b}_f~\mathbf{b}_\omega~\mathbf{c}~\mathbf{z}}_\text{robot states, $l_s$}~\vertbar~\underbrace{\boldsymbol{\mu}_0,~\cdots~\boldsymbol{\mu}_J~\rho_0~\cdots~\rho_J}_\text{features states, $l_f$}]^T
\end{eqnarray}
\normalsize
where $l_p,l_s,l_f$ are dimensions, $\mathbf{r}$ is the robocentric position of the IMU expressed in $\Bs$ (Figure~\ref{fig:coordframes}), $\boldsymbol{\upsilon}$ represents the robocentric velocity of the IMU expressed in $\Bs$, $\mathbf{q}$ is the IMU attitude represented as a map from $\Bs \rightarrow \Ws$, $\mathbf{b}_f$ represents the additive accelerometer bias expressed in $\Bs$, $\mathbf{b}_\omega$ stands for the additive gyroscope bias expressed in $\Bs$, $\mathbf{c}$ is the translational part of the IMU--cameras extrinsics expressed in $\Bs$, $\mathbf{z}$ represents the rotational part of the IMU--cameras extrinsics and is a map from $\Bs \rightarrow \Vs$, while $\boldsymbol{\mu} _j$ is the bearing vector to feature $j$ expressed in $\Vs$ and $\rho_j$ is the depth parameter of the $j^{th}$ feature such that the feature distance $d_j$ is $d(\rho_j) = 1/\rho_j$. The relevant state propagation and update steps are briefly summarized in Table~\ref{tab:roviosummary}, and are detailed in~\cite{bloesch2015robust}. Calibration of camera--IMU extrinsics takes place using the work in~\cite{furgale2013unified}. Given the estimation of the robot pose, this is then expressed on $\Ws$ the world frame and the relevant pose transformations $_{k-1}^{~~~k}\mathbf{T}_{VI}$ are available to be utilized by the proposed VI odometry--enhanced geometrically ICP algorithm.

\begin{table}[]
\centering
\caption{ROVIO State Propagation \& Filter Update Steps. }
\label{tab:roviosummary}
\begin{tabular}{|L{8.15cm}|}
\hline 
\multicolumn{1}{|c|}{State Propagation Step - Equations (3)}\\ \hline\hline


$\begin{aligned}\label{eq:lnnonspbb}
 \dot{\mathbf{r}} &= -\hat{\boldsymbol{\omega}}^{\times}\mathbf{r} + \boldsymbol{\upsilon} + \mathbf{w}_r\\[-2pt]
\dot{\boldsymbol{\upsilon}} &= -\hat{\boldsymbol{\omega}}^{\times}\boldsymbol{\upsilon} + \hat{\mathbf{f}} + \mathbf{q}^{-1}(\mathbf{g})\\[-2pt]
\dot{\mathbf{q}} &= -\mathbf{q}(\hat{\boldsymbol{\omega}})\\[-2pt]
\dot{\mathbf{b}}_f &= \mathbf{w}_{bf}\\[-2pt]
\dot{\mathbf{b}}_{\omega} &= \mathbf{w}_{bw}\\[-2pt]
\dot{\mathbf{c}} &= \mathbf{w}_c \\[-2pt]
\dot{\mathbf{z}} &= \mathbf{w}_z\\[-2pt]
\dot{\boldsymbol{\mu}}_j &= \mathbf{N}^T(\boldsymbol{\mu}_j)\hat{\boldsymbol{\omega}}_{\Vs} - \begin{bmatrix}
0 & 1\\ 
-1 & 0
\end{bmatrix}\mathbf{N}^T (\boldsymbol{\mu}_j)\frac{\hat{\boldsymbol{\upsilon}}_\Vs}{d(\rho_j)}+\mathbf{w}_{\mu,j}\\[-2pt]
\dot\rho_j &= -\boldsymbol{\mu}_j^T \hat{\boldsymbol{\upsilon}}_\Vs / d^\prime (\rho_j) + w_{\rho,j}
\end{aligned}$
\\
\horzbar\\
$\begin{aligned}
 \hat{\mathbf{f}} &= \tilde{\mathbf{f}} - \mathbf{b}_f -\mathbf{w}_f \\[-2pt]
 \hat{\boldsymbol{\omega}} &= \tilde{\boldsymbol{\omega}} - \mathbf{b}_\omega - \mathbf{w}_\omega\\[-2pt]
 \hat{\boldsymbol{\upsilon}}_\Vs &= \mathbf{z}(\boldsymbol{\upsilon} + \hat{\boldsymbol{\omega}}^{\times} \mathbf{c})\\[-2pt]
 \hat{\boldsymbol{\omega}}_{\Vs} &= \mathbf{z}(\hat{\boldsymbol{\omega}})
\end{aligned}$

          \\ \hline\hline
\multicolumn{1}{|c|}{Filter Update Step - Equations (4)} \\ \hline\hline


\scriptsize
$\begin{aligned}
 \mathbf{y}_j &= \mathbf{b}_j(\boldsymbol{\pi}(\hat{\boldsymbol{\mu}}_j)) + \mathbf{n}_j \\[-2pt]
 \mathbf{H}_j &= \mathbf{A}_j (\boldsymbol{\pi}(\hat{\boldsymbol{\mu}}_j)) \frac{d\boldsymbol{\pi}}{d\boldsymbol{\mu}}(\hat{\boldsymbol{\mu}}_j)
\end{aligned}$
\normalsize\\

By stacking the above terms for all visible features, standard EKF update step is directly performed to derive the new estimate of the robot belief for its state and the tracked features. 

                     \\ \hline\hline
\multicolumn{1}{|c|}{Notation} \\\hline\hline

$^{\times}\rightarrow$ skew symmetric matrix of a vector, 
$\tilde{\mathbf{f}}\rightarrow$ proper acceleration measurement, 
$\tilde{\boldsymbol{\omega}}\rightarrow$ rotational rate measurement, 
$\hat{\mathbf{f}}\rightarrow$ biased corrected acceleration, 
$\hat{\boldsymbol{\omega}}\rightarrow$ bias corrected rotational rate, 
$\mathbf{N}^T(\boldsymbol{\mu})\rightarrow$ projection of a $3\textrm{D}$ vector onto the $2\textrm{D}$ tangent space around the bearing vector, 
$\mathbf{g}\rightarrow$ gravity vector, 
$\mathbf{w}_\star\rightarrow$ white Gaussian noise processes, 
$\boldsymbol{\pi}(\boldsymbol{\mu})\rightarrow$ pixel coordinates of a feature, 
$\mathbf{b}_i(\boldsymbol{\pi}(\hat{\boldsymbol{\mu}}_j))\rightarrow$ a $2\textrm{D}$ linear constraint for the $j^{th}$ feature which is predicted to be visible in the current frame with bearing vector $\hat{\boldsymbol{\mu}}_j$

\\ \hline

\end{tabular}
\end{table}

\subsection{VI odometry--enhanced Geometrically Stable ICP}\label{subs:vi_icp}

Given the operation of the VI odometry, the proposed approach relies on four further key steps, namely a) the use of this odometry as a prior for the ICP, b) the selection of the overlapping segments of each two consecutive point clouds based on that prior, c) the evaluation of the expected geometric stability of the ICP solution, and d) as long as this test is successful, the execution of the ICP takes place (otherwise the system relies on the VI--odometry and the corresponding transformations). These four steps are detailed below. It is highlighted that beyond the sensor intrisics and the camera--IMU extrinsics, the system also relies on appropriate calibration of the extrinsics between the ToF depth sensor and the camera~\cite{unnikrishnan2005fast}.



\subsubsection{VI odometry prior for the ICP}\label{subs:vi_prior}

As the system performs VI odometry at update rates at least as high the ICP step, the transformation $_{k-1}^{~~~k}\mathbf{T}_{D}$ that provides the translational and rotational components $^{~~~k}_{k-1}\mathbf{t}_{D},~_{k-1}^{~~~k} \mathbf{r}_{D}$ respectively between the previous and the current point clouds $\mathbf{P}^{k-1},\mathbf{P}^{k}$ of the depth sensor are available. For the derivation of this transformation, we need to account for the extrinsics calibration of the depth sensor to the VI--system as captured from the transformation $_{VI}^{~D}\mathbf{T}$. Then it holds that $_{k-1}^{~~~k}\mathbf{T}_{D} = _{VI}^{~D}\mathbf{T} _{k-1}^{~~~k}\mathbf{T}_{VI}$, where $_{k-1}^{~~~k}\mathbf{T}_{VI}$ is the transformation of the VI--odometry. It is highlighted that the relative transformation from the VI is added to the previous ICP--derived transformation as the two processes are not necessarily aligned. 

\subsubsection{Input Point Cloud Noise Reduction}\label{subs:pcl_filter}

Employing a voxel grid filter at a fine resolution, the input point cloud is filtered for noise. For simplicity of notation, we maintain the same symbols $\mathbf{P}^{k-1}$ and $\mathbf{P}^{k}$ for the point clouds after this step. 

\subsubsection{Evaluation of Input Point Cloud Overlap}\label{subs:pcloverlap}

\begin{figure}[h!]
\centering
  \includegraphics[width=0.65\columnwidth]{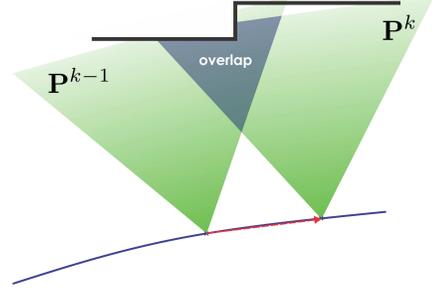}
\caption{Employed coordinate frames.  }
\label{fig:pcl_overlap}
\end{figure}

Given $_{k-1}^{~~~k}\mathbf{T}_{D}$ and the point clouds $\mathbf{P}^{k-1},\mathbf{P}^{k}$ we are able to derive $\mathbf{P}^{k-1}_{C} \subseteq \mathbf{P}^{k-1}$,~ $\mathbf{P}^{k}_{C} \subseteq \mathbf{P}^{k}$ which correspond to the subsets of $\mathbf{P}^{k-1}, \mathbf{P}^{k}$ that overlap with each other as evaluated based on the transformation prior $_{~~~k}^{k-1}\mathbf{T}_{D}= (_{k-1}^{~~~k}\mathbf{T}_{D})^{-1}$. The relevant process is visualized in Figure~\ref{fig:pcl_overlap}. Furthermore, the translated and rotated $\mathbf{P}^{k}_{C,T}$ is computed and accounts for the preliminary alignment of $\mathbf{P}^{k}_{C}$ based on $_{k-1}^{~~~k}\mathbf{T}_{D}$. Calculating the overlap has certain benefits, namely a) the operation on ICP on reduced yet equally informative data therefore enabling faster computations, b) robustify the ICP behavior against possibly erroneous data association and matching. 

%
%

\subsubsection{Evaluation of Expected Geometric Stability}\label{subs:geomstability}

With the overlapping subset of the point cloud $\mathbf{P}^{k}_{C,T}$ and $\mathbf{P}^{k-1}_{C}$ at hand, the algorithm further proceeds to the evaluation of the expected geometric stability of the ICP. As ICP is a non--linear local search algorithm, it suffers from multiple problems common to this class of optimization processes. Those include slow convergence, as well as the tendency to fall into local minima~\cite{gelfand2003geometrically,rusinkiewicz2001efficient}. Therefore, the point selection strategy and the choice of the error metric to be minimized contribute significantly into the rate of convergence and the accuracy of the resulting map and pose estimation. 

\begin{figure}[h!]
\centering
  \includegraphics[width=0.95\columnwidth]{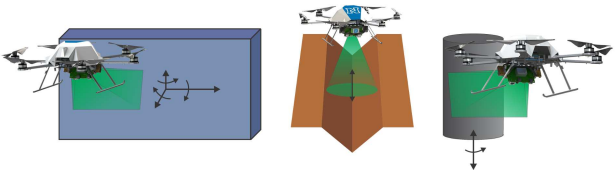}
\caption{Examples of environment geometries and trajectories that can lead to depth sensor ill--conditioned and therefore not well--constrained, unstable alignment of point cloud data. }
\label{fig:detectability}
\end{figure}

A critical question is if the (selected) points within the two consecutive point clouds can well--constrain the pose estimation problem. Our interest into this problem rises from the fact that if a set of data used in the optimization process is degenerate then one cannot expect reliable results. This is the case when a robot operates in environments and follows trajectories that lead to sensor ill--conditioning. The case of a rotorcraft vehicle using a depth sensor and flying facing a flat wall or across a symmetric canyon are obvious examples and Figure~\ref{fig:detectability} provides relevant visualizations. Essentially, this discussion is similar to that of detectability and identifiability in systems analysis and identification methods. A certain change in position of the sensor has to be able to lead to a \textit{detectable} unambiguous change on the sensor data. Furthermore, the identifiability of the translational and rotational components of the pose--to--pose transformation have to be \textit{identifiable}. Identifiability is the property which a process has to satisfy in order for precise inference to be possible. As the input data depend also on the geometry of the environment and the particular sequence of sensor positioning, we eventually have to be able to assess if the specific data can constrain the optimization process and lead to a reliable estimate of the pose--to--pose transformation. 

%
%

Towards that goal, let the $\ell$ point--pairs $\mathbf{p}_i^k,\mathbf{q}_i^k$ correspond to the closest point pairs in $\mathbf{P}^{k-1}_{C}$ and $\mathbf{P}^{k}_{C,T}$ given an initial selection in $\mathbf{P}^{k-1}_{C}$. Further let $\mathbf{n}_i^k$ denote the normals of each $\mathbf{q}_i^k$. Then the alignment error is given by:

\vspace{-1.5ex}
\small
\begin{eqnarray}\label{eqs:icp1}
 E = \sum_{i=1}^{\ell} (( _{k-1}^{~~~k}\mathbf{R}_{ICP} \mathbf{p}_i^k + _{k-1}^{~~~k}\mathbf{t}_{ICP} - \mathbf{q}_i^k)\cdot \mathbf{n}_i^k)^2
\end{eqnarray}
\normalsize
where $_{k-1}^{~~~k}\mathbf{R}_{ICP},~ _{k-1}^{~~~k}\mathbf{t}_{ICP}$ are the rotational and translational components of the transformation to be estimated. As the prealignment step based on $_{k-1}^{~~~k}\mathbf{T}_{D}$ has been conducted, we can assume that the rotation that minimizes $E$ is small. Therefore the problem can be linearized~\cite{gelfand2003geometrically} and re--casted to that of the identification of a $6$--vector $[_{k-1}^{~~~k}\mathbf{r}_{ICP}^T,~ _{k-1}^{~~~k} \mathbf{t}_{ICP}^T]$ based on the minimization of:

\vspace{-1.5ex}
\scriptsize
\begin{eqnarray}\label{eqs:icp2}
 E = \sum_{i}^{\ell} (( \mathbf{p}_i^k - \mathbf{q}_i^k )\cdot \mathbf{n}_i^k + _{k-1}^{~~~k}\mathbf{r}_{ICP} \cdot ( \mathbf{p}_i^k \times  \mathbf{n}_i^k) + _{k-1}^{~~~k}\mathbf{t}_{ICP} \cdot \mathbf{n}_i^k ) ^2
\end{eqnarray}
\normalsize
where the terms of this equation can be considered as imaginary ``forces'' in the direction of $\mathbf{n}_i^k$ and ``torques'' around the axis $\mathbf{p}_i^k \times  \mathbf{n}_i^k$. 

By taking the partial derivatives of Eq.~\ref{eqs:icp2}, a linear system of the form $\mathbf{C}\mathbf{x}=\mathbf{b}$ is derived, where $\mathbf{x}$ is the $6\times 1$ vector of transformation parameters, $\mathbf{b}$ is the residual vector, and $\mathbf{C}$ is a $6\times 6$ covariance matrix:

\vspace{-1.5ex}
\scriptsize
\begin{eqnarray}\label{eqs:icp3}
\mathbf{C} = \mathbf{F} \mathbf{F}^T =  \begin{bmatrix}
\mathbf{p}_1^k \times \mathbf{n}_1^k & ... & \mathbf{p}_1^\ell \times \mathbf{n}_1^\ell \\ 
\mathbf{n}_1^k  & ... & \mathbf{n}_\ell^k 
\end{bmatrix}\begin{bmatrix}
(\mathbf{p}_1^k \times \mathbf{n}_1^k)^T & (\mathbf{n}_\ell^k)^T\\ 
... & ...\\ 
(\mathbf{p}_\ell^k \times \mathbf{n}_\ell^k)^T & (\mathbf{n}_\ell^k)
\end{bmatrix}
\end{eqnarray}
\normalsize
Within that, the matrix $\mathbf{C}$ encodes how sensitive the alignment is when $\mathbf{P}^{k-1}_{C}$ is moved from its optimum alignment with $\mathbf{P}^{k}_{C,T}$ by a transformation $[\Delta _{k-1}^{~~~k}\mathbf{r}_{ICP}^T, \Delta _{k-1}^{~~~k} \mathbf{t}_{ICP}^T]$: 

\vspace{-1.5ex}
\small
\begin{eqnarray}\label{eqs:icp4}
\begin{bmatrix}
\Delta _{k-1}^{~~~k}\mathbf{r}_{ICP}^T & \Delta _{k-1}^{~~~k} \mathbf{t}_{ICP}^T
\end{bmatrix} \mathbf{C}  
 \begin{bmatrix}
 \Delta _{k-1}^{~~~k}\mathbf{r}_{ICP}
\\ 
\Delta _{k-1}^{~~~k} \mathbf{t}_{ICP}
\end{bmatrix}
\end{eqnarray}
\normalsize
As shown in~\cite{gelfand2003geometrically}, the transformations for which this increase in error is relatively small, correspond to the directions where the input depth data can slide relative to each other. 

Given this analysis, and based on the fact that certain types of geometry will lead to a covariance matrix that is close to the singularity point, not all input data can lead to a well-constrained solution. Therefore in this work we utilize a metric of the expected stability of the ICP solution that detects if the input depth data presents any rotational or translational instability. As in~\cite{gelfand2003geometrically}, the relevant unconstrained transformations can be identified by expressing $\mathbf{C}$ in terms of its eigenvalues and eigenvectors. Let $\mathbf{x}_1,...,\mathbf{x}_6$ be the eigenvectors of $\mathbf{C}$ and $\mathbf{\lambda}_1 \ge ... \ge \mathbf{\lambda}_6$ the ordered set of corresponding eigenvalues. If any of the $\lambda_j$ are small compared to $\lambda_1$, then the associated eigenvector corresponds to a sliding direction and relates to a transformation that can disturb the two input point clouds from their optimum alignment with only a small increase in error (less stable optimization). Therefore, the employed \textit{metric of stability} is the condition number ($c$)~\cite{gelfand2003geometrically}: 

\vspace{-1.5ex}
\small
\begin{eqnarray}
 c = \frac{\lambda_1}{\lambda_6}
\end{eqnarray}
\normalsize
As $c$ gives an estimate of the geometric stability of the ICP solution, within our approach we utilize it for two purposes. First, we evaluate it as a guidance for the selection of a high--performance sampling method for appropriate sampling of points within the point cloud data that are to be associated. Second, we define a threshold $c_{thres}$ above which ICP will not be attempted and the system will purely rely on VI odometry and the associated transformation $_{k-1}^{~~~k}\mathbf{T}_{D}$. The calculation of the condition number relies on the methodology employed to select and sample data from the input point clouds. This is detailed below.

\subsubsection{ICP Algorithm}\label{subs:roboticp}

When a good prior exists, the ICP optimization may be executed. This process follows multiple steps, namely that of the selection of some set of points in the two point clouds, matching between these points, weighting of the derived matched pairs, rejection of outliers, definition of the error metric and solution of the associated minimization problem. For our implementation we make use of the Point Cloud Library (PCL)~\cite{rusu20113d}.


Of particular interest, is the strategy to select the sets of points to be matched and a variety of relevant methods exist. Those include a) the use of all available points~\cite{Besl92}, b) uniform subsamping of the available points~\cite{turk1994zippered}, c) random sampling (with a different sample of points for each iteration)~\cite{masuda1996registration}, d) selection of points based on trying to equally constraint the eigenvectors in $\mathbf{C}$~\cite{gelfand2003geometrically}, e) selection of points such that the distribution of normals among them is as large as possible~\cite{rusinkiewicz2001efficient}, and more~\cite{pomerleau2013comparing}.

In the literature, it is often discussed that high variability regarding the performance of the several data sampling methods is observed~\cite{rusinkiewicz2001efficient,pomerleau2013comparing}. This is especially the case in natural, unstructured and information-deprived environments where no particular ``type'' of geometry is dominant and no trivial cases are encountered. Our evaluation between the methods mentioned before and detailed in~\cite{Besl92,turk1994zippered,masuda1996registration,gelfand2003geometrically,rusinkiewicz2001efficient} indicated that the selection of points such that the distribution of normals among them is as large as possible (\textit{``normals--based''}) provided an overall superior performance for environments such as rooms and other man--made facilities relevant to inspection applications, while also leading to a reliable calculation of the condition number and its coherence with the actual geometric stability of the ICP optimization step. Therefore this method is employed. 

Based on the above, given the overlapping point cloud data $\mathbf{P}^{k-1}_{C}$ and $\mathbf{P}^{k}_{C,T}$, the algorithm proceeds with the selection of data based on the normals--based method, evaluates the condition number and as long as this is below the threshold $c_{thres}$, it moves forward with the execution of the remaining steps of the ICP in order to estimate the transformation $_{k-1}^{~~~k}\mathbf{T}_{ICP}$. When the condition number is above $c_{thres}$, the overall framework relies on the VI odometry based transformation $_{k-1}^{~~~k}\mathbf{T}_{D}$. For the steps of points matching, weighting, outlier rejection, and the minimization of the error metric we rely on the PCL library implementations employing the point--to--plane error metric, RANSAC outlier rejection, uniform weighting and Singular Value Decomposition--based transformation estimation. 

\section{EXPLORATION PLANNER SYNOPSIS}\label{sec:planner}
\begin{figure}[h!]
\centering
  \includegraphics[width=0.99\columnwidth]{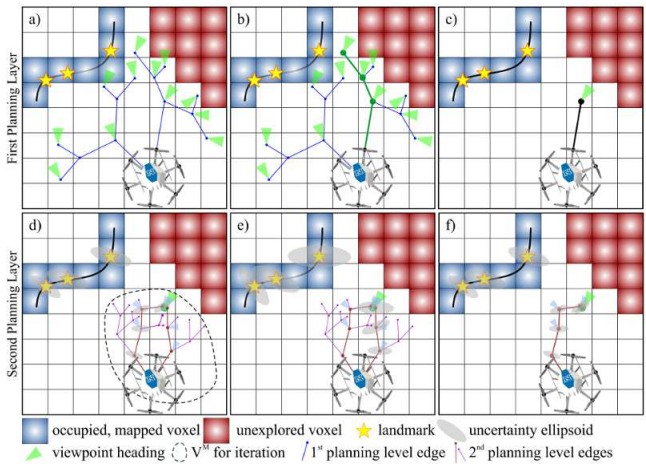}
\caption{2D representation of the two--steps uncertainty--aware exploration and mapping planner. The first planning layer samples the path with the maximum exploration gain. The viewpoint configuration of the first vertex of this path becomes the reference to the second planning layer. Then this step, samples admissible paths that arrive to this configuration, performs belief propagation along the tree edges, and selects the one that provides minimum uncertainty over the robot pose and tracked landmarks.  }
\label{fig:rhem_steps}
\end{figure}

Beyond the evaluation of the proposed pose estimation and mapping approach, this work further contributes with the verification of its use as part of autonomous exploration missions. For the purposes of this task, our previous work on real--time uncertainty--aware Receding Horizon Exploration and Mapping (RHEM) planner~\cite{RHEM_ICRA_2017} is employed. The RHEM planner relies on a two--step, receding horizon, belief space--based approach. At first, in an online computed random tree, the algorithm identifies the branch that optimizes the amount of new space expected to be explored. The first viewpoint configuration of this branch is selected, but the path towards it is decided through a second planning step. Within that, a new tree is sampled, admissible branches arriving at the reference viewpoint are found and the robot belief about its state and the tracked landmakrs is propagated. As system state the concatenation of the robot states and tracked landmarks (visual features) is considered as in Eq.~\ref{eq:roviostate}. Then, the branch that minimizes the localization uncertainty, as factorized using the D--optimality (D--opt)~\cite{carrillo2012comparison} of the pose and landmarks covariance is selected. The corresponding path is conducted by the robot and the process is iteratively repeated. Figure~\ref{fig:rhem_steps} illustrates the basic steps of this planner. 

%
%

The planner assumes a volumetric, incrementally built map $\Ms$ as in~\cite{hornung13auro} which is reconstructed using the robot's perception unit and further annotates every voxel $m \in \Ms$ with a probability indicating how certain the mapping is about it being occupied. For the first planning step, a random tree (e.g. RRT~\cite{RRT}) $\mathbb{T}^E$ is sampled in the position--heading space and its vertices are annotated with the cummulative ``exploration gain'' (given the sensor model and the current map $\Ms$) alongside with a ``reobservation gain'' expressing the importance of reobserving voxels with low mapping certainty. For the sampled state configuration $\xi_{k}$, the exploration gain at the vertex $n_k^E$ is defined as: 

\vspace{-1.5ex}
\scriptsize
\begin{eqnarray}
 \mathbf{ExplorationGain}(n^E_k) = \mathbf{ExplorationGain}(n_{k-1}^E) + \\ \nonumber \mathbf{VisibleVolume}(\Ms,\xi_k)\exp(-\lambda c(\sigma_{k-1,k}^E)) +\\ \nonumber \mathbf{ReobservationGain}(\Ms,\Ps,\xi_k) \exp(-\lambda c(\sigma_{k-1,k}^E))
\end{eqnarray}
\normalsize
where $\sigma_{k-1,k}^E$ denotes the path from the sampled configuration $\xi_{k-1}$ to $\xi_{k}$, $c(\sigma_{k-1,k}^E)$ is the length of the path, and $\lambda$ is a parameter to penalize long paths. At the end of this step, the path that maximally explores is identified and its first viewpoint configuration $\xi_{RH}$ is returned as detailed in~\cite{RHEM_ICRA_2017}.

Given $\xi_{RH}$, the second planning step ensures that the way to visit this ``next--best--view'' is through a path that minimizes the localization uncertainty of the robot. To achieve this task, it samples a new random tree $\mathbb{T}^M$ that searches multiple branches to arrive to $\xi_{RH}$ and evaluates which one leads to minimized localization uncertainty (see Figure~\ref{fig:rhem_steps}). For this step, the visual--inertial localization of the robot summarized in Section~\ref{subs:vislam} is considered and the state propagation and update steps as presented and detailed in~\cite{bloesch2015robust,RHEM_ICRA_2017} are conducted. It is noted that the state vector to be propagated is that in Eq.~\ref{eq:roviostate}. The belief propagation steps are executed for the branches $\sigma^M_\alpha$ of the tree that arrive close to $\xi_{RH}$. For all $\sigma^M_\alpha$ the planner evaluates the ``belief gain'' at the final configuration. This is calculated as the D--opt of the subset of the covariance matrix $\mathbf{\Sigma}_{p,f}$ that refers to the robot pose and landmarks:

\vspace{-1.5ex}
\small
\begin{eqnarray}
 & \mathbf{BeliefGain}(\sigma^M_\alpha) = \nonumber \\
 & D_{opt}(\sigma^M_\alpha) = \exp(\log([\det(\mathbf{\Sigma}_{p,f}(\sigma^M)]^{1/(l_p + l_f)}))
\end{eqnarray}
\normalsize
Through this process, the path that leads to minimized localization uncertainty, while arriving at the viewpoint provided by the exploration planning step, is achieved.


\section{EXPERIMENTAL EVALUATION}\label{sec:experiments}
For the goal of detailed evaluation of the proposed VI odometry--enhanced geometrically stable ICP, a set of studies, including handheld datasets, as well as automated and autonomous robotic missions were conducted. Within those, the handheld experiments were conducted as preliminary verification. Automated inspection experiments with predefined inspection routes were conducted to evaluate the mapping performance given a detailed set of viewpoints, while autonomous exploration and mapping experiments were performed to evaluate the final mapping result for viewpoints and trajectories planned by the robot given no prior information about its environment. For all the experiments, the same sensing unit detailed in Section~\ref{subs:perceptionunit} was employed. For the experiments involving an aerial robot, a basic overview of the utilized configuration is provided in Section~\ref{subs:aerialrobot}.

\subsection{Perception Unit}\label{subs:perceptionunit}

A Visual-Inertial-Depth perception unit was employed for the purposes of this study. The VI--sensing modalities were facilitated using an electronically synchronized rolling shutter camera (StereoLabs, ZED) that is software synchronized with a low--cost IMU (UM--7). The employed ToF depth sensor was the PMD Picoflexx integrating the IRS1145C Infineon\textregistered~3D Image Sensor. The data from all modalities are processed onboard an Intel NUC5i7RYH with the VI odometry running at $20\textrm{Hz}$ and ICP executed at $1\textrm{Hz}$.

\subsection{Aerial Robotic Platform}\label{subs:aerialrobot}

\begin{figure}[h!]
\centering
  \includegraphics[width=0.99\columnwidth]{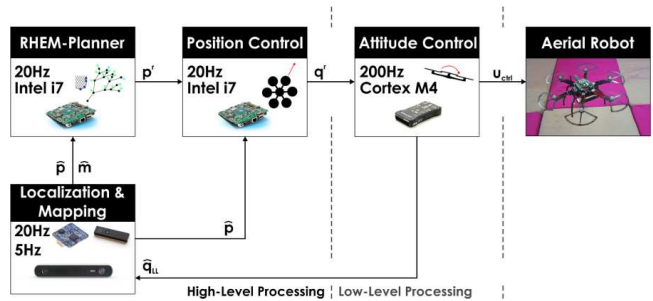}
\caption{Overview of the key robot functionalities.}
\label{fig:hardware}
\end{figure}

A custom--built hexarotor platform is employed and has a weight of $2.6\textrm{kg}$. The system is relying on a Pixhawk--autopilot for its attitude control, while further integrating the perception unit mentioned above alongside the NUC5i7RYH. A block diagram overview is depicted in Figure~\ref{fig:hardware}. As shown, all the localization and mapping, planning and position control loops are running on the NUC5i7RYH. For the position control task, a linear model predictive controller~\cite{mpc_rosbookchapter} is utilized. The complete set of high--level tasks run with the support of the Robot Operating System (ROS).

%
%

\subsection{Implementation Details}

This section provides further implementation details and insights to our approach. Table~\ref{tab:icp_pars} details the tuning parameters for the ICP step, where $d_{corr}^{\max},~t_{RNSC}^{reject},~e_{T},M_{iter}^{\max},~e_{E}^{f},~c_{thres}$ stand for the ICP maximum corresponence distance, ICP RANSAC outlier rejection, ICP transformation epsilon, the ICP maximum number of iterations, the ICP euclidean distance epsilon, and the condition number threshold. 

\begin{figure}[h!]
\centering
  \includegraphics[width=0.85\columnwidth]{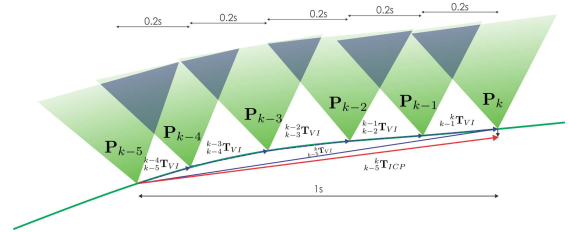}
\caption{Point clouds derived at $5\textrm{Hz}$ are merged in their neighborhood and used altogether for the estimation of the ICP transformation of the robot motion at $1\textrm{Hz}$.  }
\label{fig:merging}
\end{figure}

At the same time, it should be noted that although the employed depth sensor can provide data at high update rates, the computational cost of ICP challenges our ability to make use of an increased frame rate as long as we want to allow for sufficient iterations. A balance between frame rate, data quality and number of iterations has to be found. In that sense, it was identified that an update rate of $1\textrm{Hz}$ performs sufficiently well and allows to keep the overall computational load sufficiently low. However, if one was to neglect the consideration of computational load, use of higher update rate would be the natural choice. As a hybrid approach, we propose to further exploit the VI odometry information in a way that will allow us to run ICP at $1\textrm{Hz}$ while using all the data coming from the sensor set to deliver point clouds at $5\textrm{Hz}$. In that sense, from the last point cloud and using all the $5$ consecutive point clouds we propose at a first step to find the overlap from all the point clouds to their neighboring ones (exploiting the VI priors), and subsequently conduct one ICP step aiming to robustly identify the robot pose transformation that optimally explains its motion based on this combined point cloud set. Figure~\ref{fig:merging} illustrates this procedure. It is noted that this methodology is viable as long as small translational and rotational deviations take place within the frame duration ($0.2\textrm{s}$) which is the case given the small velocities employed by the robot.

%
%

\begin{table}[]
\centering
\renewcommand{\arraystretch}{1.35}
\caption{Key ICP algorithm parameters\label{tab:icp_pars}}
\begin{tabular}{|l|l|}
\hline
\multicolumn{1}{|c|}{\textbf{Parameter}} & \multicolumn{1}{c|}{\textbf{Value}} \\ \hline
$d_{corr}^{\max}$                        & $0.01$                             \\ \hline
$t_{RNSC}^{reject}$                      & $0.01$                              \\ \hline
$e_{T}$                                  & $1e^{-8}$                              \\ \hline
$M_{iter}^{\max}$                        & $1000$                              \\ \hline
$e_{E}^{f}$                              & $0.005$                             \\ \hline
$c_{thres}$                               & $15$                                 \\ \hline
\end{tabular}
\end{table}

\subsection{Handheld Evaluation}

\begin{figure}[h!]
\centering
  \includegraphics[width=0.99\columnwidth]{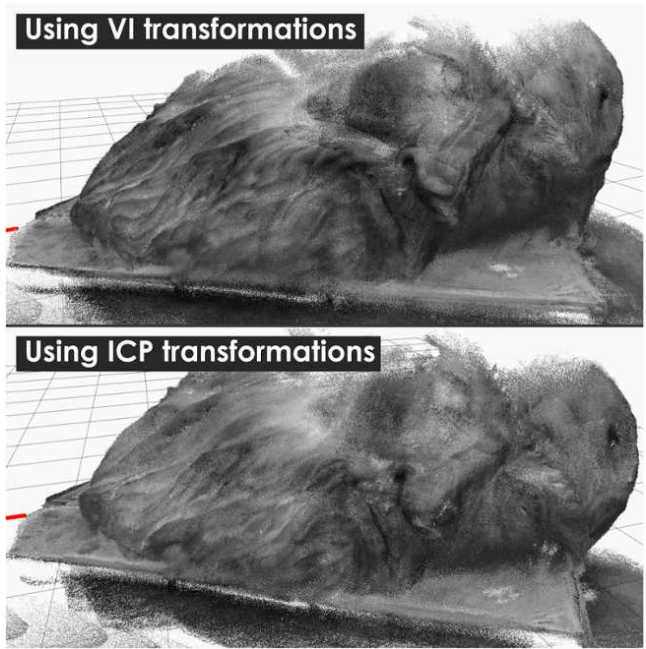}
\caption{Handheld mapping of a the ``Leonardo da Vinci head'' statue at the university campus with an approximate length. This tests was used to verify the robustness of the proposed visual--inertial odometry--enhanced ICP approach. The two point clouds, using the transformations from the VI odometry and those from ICP are comparable with one notable geometric fix in the case of the ICP being the more representative shape of the nose.  }
\label{fig:handheld}
\end{figure}

For a preliminary evaluation and in order to verify that the proposed visual--inertial odometry--enhanced geometrically stable ICP performs robustly, a handheld test was conducted. This refers to the $3\textrm{D}$ reconstruction of the ``Leonardo da Vinci head'' statue at the university campus with an approximate length of $7\textrm{m}$. The relevant result is shown in Figure~\ref{fig:handheld} and verifies that the method is reliable.  

%
%

\subsection{Evaluation in Automated Robotic Inspection}

\begin{figure}[h!]
\centering
  \includegraphics[width=0.99\columnwidth]{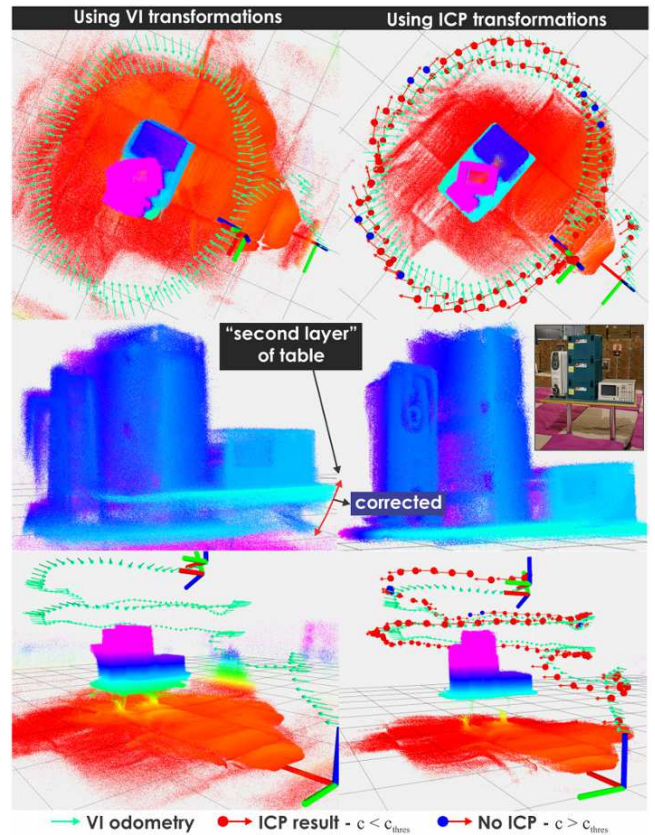}
\caption{Mapping results following an automated inspection route ensuring the coverage of a geometry of objects of interest. As shown, the ICP--derived transformations are superior to those coming from the VI odometry alone. This results to the removal of a ``double laying'' effect regarding the table of the structure. At several times, the condition number indicated that not performing the ICP step and relying on the VI transformations was the best choice and ensured the robustness of the approach. }
\label{fig:automatedroute}
\end{figure}

In this experiment, the aerial robot is instructed to follow a fixed set of waypoints designed to provide full coverage of a geometry of objects of interest. Figure~\ref{fig:automatedroute} presents the derived results. As shown, the geometric stability check indicated multiple times the need to rely on the VI odometry transformations in order to ensure robustness. Furthermore, due to the ICP optimization step, the derived map is of high quality compared to a map created by merging the point clouds using the VI odometry. 

%
%

\subsection{Evaluation in Autonomous Robotic Exploration}

\begin{figure}[h!]
\centering
  \includegraphics[width=0.99\columnwidth]{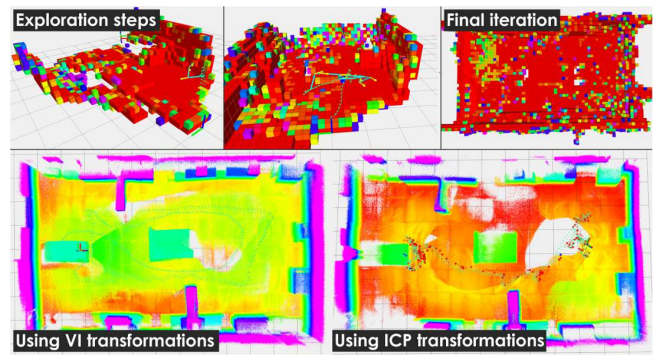}
\caption{Exploration steps of the uncertainty--aware receding horizon exploration and mapping planner alongside with mapping results using the VI odometry--provided transformations and those after the geometrically stable ICP step. In this long mission we demonstrate that the proposed system robustly provides updated transformations without a breaking point as can be often the case for ICP implementations that do not account for the geometric stability check and do not exploit a reliable prior such as the one provided from the VI odometry. }
\label{fig:autonomousexploration}
\end{figure}

The autonomous exploration scenario refers to the mapping of an indoors room with dimensions $12\times6.5\times2\textrm{m}$. Using $300$ boxes with size $0.4\times0.3\times0.3\textrm{m}$, vertical and T--shaped walls, as well as other structural elements are created to complexify the robot exploration and mapping mission. For this task the RHEM planner summarized in Section~\ref{sec:planner} is employed and the robot is constrained to fly with a maximum velocity of $v_{\max}=0.75\textrm{m/s}$ and maximum yaw rate $\dot{\psi}_{\max}=\pi/4\textrm{rad/s}$. Figure~\ref{fig:autonomousexploration} presents the derived mapping result. The ICP step improves the quality of the overall map, although due to the size of the environment specific improvements are not equally distinguishable as in the previous case. 


%
%

\begin{figure}[h!]
\centering
  \includegraphics[width=0.99\columnwidth]{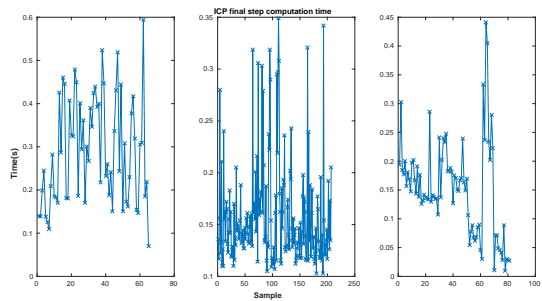}
\caption{ICP computational times for all three sets presented in this work (from left to right: handheld, automated, autonomous). As shown, in all cases and within all iterations, the computational time is below the available $1\textrm{s}$, while running onboard the robotic system alongside the remaining tasks. }
\label{fig:icptimes}
\end{figure}

Finally, Figure~\ref{fig:icptimes} verifies that the ICP step required computational time less than the available $1\textrm{s}$ for all the iterations of all the three presented datasets.

%
%

\section{CONCLUSIONS}\label{sec:concl}

An approach for visual--inertial odometry--enhanced geometrically stable ICP for aerial robotic mapping applications was presented in this paper. Odometry priors from the visual--inertial localization module are utilized to support the ICP optimization step in the sense of computing overlapping regions between the input depth point clouds, providing sufficiently robust priors and maintaining the system operation when depth sensor data are ill--conditioned. A check of geometric stability allows us to evaluate the capability of ICP to reliably estimate the pose transformation. It is only then that the ICP step is conducted and the derived map is optimized. Extended evaluation in handheld, as well as automated and autonomous aerial robotic inspection missions verified the robustness and efficiency of the proposed approach.

\bibliographystyle{IEEEtran}
\bibliography{./BIB/QMAP_IROS2017_BIBitems}

\end{document}